\documentclass[letterpaper, 10 pt, conference]{ieeeconf}

\IEEEoverridecommandlockouts
\usepackage[utf8]{inputenc}
\usepackage[T1]{fontenc}
\usepackage{amsmath,amssymb,booktabs,array,calc,color,ifthen,capt-of,textcomp}
\usepackage{graphicx}
\usepackage{url}
\usepackage{listings}
\usepackage{xcolor}

\providecommand{\tightlist}{%
  \setlength{\itemsep}{0pt}\setlength{\parskip}{0pt}}
\providecommand{\passthrough}[1]{#1}
\title{\LARGE \bf
MawForge: Memory-Bounded Expert Materialization for Local Mixture-of-Experts Inference
}

\author{Craig Opie, Holocron Security, Inc.}

\begin{document}

\maketitle
\thispagestyle{empty}
\pagestyle{empty}

\begin{abstract}
Sparse Mixture-of-Experts (MoE) language models separate total parameter count from per-token active computation, but local inference systems often still require the full model, key-value cache, runtime buffers, and operating-system headroom to fit in fast memory. MawForge tests a different systems hypothesis: local MoE serving can be made practical on constrained unified-memory machines by storing the full model on disk, keeping common tensors resident, and materializing routed expert tensors into a bounded execution cache on demand.

This article presents MawForge\textquotesingle s split-pack architecture, budget model, native GGUF serving path, and completed local validation on a MacBook Pro M5 Pro with 24 GB unified memory under an 18 GiB explicit serving target. The validation matrix contains 600 non-speculative MawForge rows across three model profiles, five cache settings, two context lengths, four prompt classes, and five repetitions per cell. Static planning accepted 27 of 30 unique model/context/cache cells and rejected 3 cells before execution; all 540 statically feasible generation rows returned valid 96-token completions without triggering the memory guard. The results show that MawForge can serve large GGUF MoE models within a bounded local memory envelope, including a 34 GB Qwen3.6 35B A3B Q8\_0 profile and a 25 GB Gemma 4 26B A4B Q8\_0 profile. They also show that expert-cache size is non-monotonic: larger caches consistently improve hit rate and reduce materialized bytes, but they can reduce throughput by increasing host memory pressure. Qwen Q8 favored the smallest tested cache, 15\%, at both 4K and 32K contexts; Gemma Q8 favored 35\% among feasible points; Qwen Q4 shifted with context and prompt class. A focused speculative-decoding addendum found that Gemma MTP speculation accepted more draft tokens but reduced throughput and increased expert materialization, indicating that dense-model speculation assumptions do not transfer directly to split-pack MoE serving.

The central finding is that MawForge is effective as a bounded execution mechanism and measurement substrate for local MoE inference, but not as a cache-maximization policy. Performance depends on balancing expert reuse against resident footprint, KV-cache size, quantization, route locality, and macOS memory pressure.
\end{abstract}


\section{Introduction}\label{introduction}

Large language model deployment is usually governed by a hard residency constraint: model weights must fit in fast memory with additional headroom for key-value (KV) cache, runtime allocations, accelerator buffers, and operating-system pressure. This constraint is especially sharp on consumer and workstation systems with unified memory because the CPU, GPU, kernel, display stack, filesystem cache, and inference runtime draw from the same physical pool.

Sparse MoE models relax the compute side of this constraint. A token does not activate every expert; it activates a routed subset. However, most local inference paths still treat model residency as a whole-file loading problem. The full GGUF is mapped or loaded, and failure occurs when the aggregate model, KV cache, and runtime memory exceed what the machine can absorb. Sparse activation does not automatically become sparse residency.

MawForge reframes the problem as a memory hierarchy. The full model remains a durable disk object. Common tensors remain resident. Routed expert tensors are split into per-layer, per-expert payloads and copied into bounded in-memory slots only when the router demands them. The resulting question is not simply "does the full model fit?" but "how much expert cache is enough to capture route locality without crossing the host memory-pressure boundary?"

This distinction matters. On the test machine, a direct non-MawForge load of Gemma 4 26B A4B Q8\_0 at 32K context crossed a 98\% system-used safety guard and was terminated before producing route output. The MawForge split-pack path completed every statically feasible Gemma Q8 32K validation row under an 18 GiB serving target. The finding does not prove that no other direct configuration could run; it demonstrates that the tested full-GGUF Metal path was unsafe or impractical on the same machine while MawForge\textquotesingle s bounded path remained controllable.

This article makes four contributions:

\begin{enumerate}
\def\labelenumi{\arabic{enumi}.}
\tightlist
\item
  It describes MawForge\textquotesingle s split-pack and runtime materialization architecture for local GGUF MoE serving.
\item
  It formulates local sparse-expert inference as a bounded cache-placement problem with explicit budget and throughput tradeoffs.
\item
  It reports a completed 600-row validation matrix on constrained Apple Silicon hardware, including static budget rejections and guarded generation completions.
\item
  It analyzes failure modes that are not visible from cache hit rate alone, including high-cache latency collapse and speculative decoding regression.
\end{enumerate}

The work is intentionally narrow in its empirical claims. It evaluates one current machine, three model profiles, two context lengths, four prompt classes, and short fixed completions. Within that boundary, the matrix is complete for the stated non-speculative MawForge protocol.

\section{Research Questions}\label{research-questions}

The evaluation addresses five research questions.

\textbf{RQ1: Feasibility.} Can MawForge serve large GGUF MoE models on a 24 GB unified-memory laptop under an 18 GiB explicit serving budget?

\textbf{RQ2: Cache behavior.} How does expert-cache percentage affect decode throughput, TTFT, cache hit rate, materialized expert bytes, and memory pressure?

\textbf{RQ3: Quantization and context.} Does changing quantization or context length change the feasible cache region and the best operating point?

\textbf{RQ4: Speculation.} Does MTP speculative decoding improve throughput for MawForge split-pack serving?

\textbf{RQ5: Measurement discipline.} What claim boundaries are justified by the completed validation evidence, and which claims require further measurement?

\section{Background and Related Work}\label{background-and-related-work}

MoE models use conditional computation to increase total parameter count while activating only a subset of parameters for each token. GShard demonstrated large-scale conditional computation and automatic sharding for sparsely gated models~\cite{lepikhin2020gshard}. Switch Transformer simplified routing and showed that sparse activation can scale model capacity while maintaining a smaller active computation path~\cite{fedus2022switch}.

Serving research has also emphasized memory management as a first-class systems problem. vLLM\textquotesingle s PagedAttention treats KV-cache allocation as a paging problem and demonstrates that memory layout and allocation policy can dominate serving throughput~\cite{kwon2023pagedattention}. MawForge applies a related systems perspective to a different object: routed expert weights rather than attention KV pages.

Speculative decoding proposes draft tokens with a smaller or cheaper model and verifies them with the target model to reduce serial decoding latency~\cite{leviathan2023speculative}. For dense models, the primary question is whether accepted draft tokens offset draft and verification overhead. For split-pack MoE serving, there is an additional question: whether speculation expands the union of routed experts touched per step enough to increase materialization cost and cache churn.

GGUF is the llama.cpp model format used for many quantized local models~\cite{ggufformat}. MawForge\textquotesingle s current most complete serving path targets GGUF MoE models through a native llama.cpp-derived worker that uses MawForge split packs rather than requiring fully resident expert tensors.

\section{MawForge Design}\label{mawforge-design}

\subsection{System Thesis}\label{system-thesis}

MawForge is built around the following thesis:

\begin{quote}
Sparse expert inference should be served from a bounded expert materialization cache rather than requiring all routed expert tensors to be resident.
\end{quote}

The thesis does not imply that disk access is free or that sparse models are automatically fast. It implies that a local inference system should expose the expert residency tradeoff directly and should make budget feasibility measurable before runtime load.

\subsection{Split-Pack Representation}\label{split-pack-representation}

The packer reads a GGUF model and separates tensors into common tensors and routed expert tensors. Common tensors are emitted in resident form. Expert tensors are split into deterministic per-layer, per-expert blocks. Each block records tensor kind, layer, expert id, source offsets, byte counts, topology metadata, and digests. The block payloads are written to \passthrough{\lstinline!experts.pack!}, and the index is written to \passthrough{\lstinline!experts.index!}.

The split-pack design has two practical consequences. First, the runtime can address an expert block without scanning or copying unrelated model weights. Second, the pack index becomes a reproducibility artifact: a validation row can bind a model profile to a pack path and pack digest rather than relying only on a human-readable model name.

\subsection{Native Serving Path}\label{native-serving-path}

The serving worker loads common tensors and allocates compact per-layer slot tensors for routed experts. Slot count is computed from the requested cache bytes, routed layer count, per-slot expert block size, active experts per token, and experts per layer. Architectures with fused feed-forward tensors use fused slot tensors; architectures with separate gate, up, and down tensors allocate corresponding slot tensors. On a cache miss, the runtime materializes the requested expert block into an available slot and updates route-to-slot metadata.

The worker exposes an OpenAI-compatible local endpoint through \passthrough{\lstinline!mawforge serve!}. The CLI reports TTFT, decode rate, expert-cache bytes, KV-cache bytes, resident common bytes, hit rate, materialized expert bytes, speculative counters, and budget provenance.

\subsection{Static Budget Check}\label{static-budget-check}

MawForge performs a static planning check before serving. Let \passthrough{\lstinline!C!} be resident common tensor bytes, \passthrough{\lstinline!E(p)!} be the expert-cache budget at cache percentage \passthrough{\lstinline!p!}, \passthrough{\lstinline!K(n)!} be KV-cache bytes at context length \passthrough{\lstinline!n!}, and \passthrough{\lstinline!O!} be runtime overhead. The static lower bound is:

\begin{lstlisting}
L(p, n) = C + E(p) + K(n)
\end{lstlisting}

The runtime footprint is:

\begin{lstlisting}
F(p, n) = C + E(p) + K(n) + O
\end{lstlisting}

Static planning rejects rows where the lower bound already exceeds the requested serving target. Runtime load then verifies allocator behavior under the same budget. This approach intentionally treats rejection as evidence: a rejected row is not a failed benchmark if the rejection proves that the requested cache/context point is outside the configured budget.

\subsection{Throughput Model}\label{throughput-model}

The cache tradeoff is non-monotonic. A simplified throughput relation is:

\begin{lstlisting}
T(p) = generated_tokens / (compute_time + materialization_time(p) + pressure_penalty(F))
\end{lstlisting}

Increasing \passthrough{\lstinline!p!} usually reduces \passthrough{\lstinline!materialization\_time(p)!} by improving route locality and hit rate. It also increases \passthrough{\lstinline!F!}. When the larger cache pushes the host into compression, paging, or lower effective memory bandwidth, \passthrough{\lstinline!pressure\_penalty(F)!} can dominate. Under that regime, cache hit rate improves while decode throughput worsens.

Speculative decoding adds another term:

\begin{lstlisting}
T_spec = generated_tokens /
         (target_time - accepted_target_work_saved
          + draft_overhead + verification_overhead
          + extra_materialization + state_overhead)
\end{lstlisting}

The critical MawForge-specific term is \passthrough{\lstinline!extra\_materialization!}. Speculative lookahead can touch a larger union of experts before route locality can amortize the cache, especially when accepted draft tokens are few in absolute number.

\section{Experimental Method}\label{experimental-method}

\subsection{Hardware and Budget}\label{hardware-and-budget}

All primary validation evidence was collected on a MacBook Pro M5 Pro with 24 GB unified memory on June 8-9, 2026 HST. The primary MawForge serving target was 18 GiB. The validation runner used memory guards and terminated direct or MawForge runs that crossed configured safety thresholds. The primary non-speculative generation rows used 96 maximum completion tokens.

\subsection{Model Profiles}\label{model-profiles}

The validation matrix used three model profiles.

\begin{table*}[t]
\centering
\scriptsize
\setlength{\tabcolsep}{3pt}
\begin{tabular}{@{}lrl@{}}
\toprule
Profile & Local size & Cache settings \\
\midrule
Gemma 4 26B A4B Q8\_0 & 25 GB & 15\%, 25\%, 35\%, 50\%, 65\% \\
Qwen3.6 35B A3B Q8\_0 & 34 GB & 15\%, 20\%, 25\%, 35\%, 45\% \\
Qwen3.6 35B A3B Q4\_K\_M & 20 GB & 15\%, 25\%, 35\%, 65\%, 90\% \\
\bottomrule
\end{tabular}
\end{table*}

The matrix evaluated two context lengths, 4K and 32K, and four prompt classes: code, summarization, long-form writing, and reasoning. Each model/context/cache/prompt cell used five repetitions when statically feasible.

\subsection{Validation Matrix}\label{validation-matrix}

The primary validation matrix is:

\begin{lstlisting}
3 model profiles * 5 cache settings * 2 contexts * 4 prompt classes * 5 repetitions = 600 rows
\end{lstlisting}

The static planning stage expanded the full 600-row manifest and executed the 30 unique model/context/cache budget checks. Of those 30 static checks, 27 were feasible and 3 were rejected under the 18 GiB budget. Each rejected static cell corresponds to 4 prompt classes and 5 repetitions, producing 60 valid static rejection rows. The remaining 540 rows were executed as guarded generation runs.

\begin{table*}[t]
\centering
\scriptsize
\setlength{\tabcolsep}{3pt}
\begin{tabular}{@{}lr@{}}
\toprule
Validation component & Count \\
\midrule
Matrix rows & 600 \\
Unique static plan rows & 30 \\
Feasible static plan rows & 27 \\
Static budget rejections & 3 unique cells, 60 matrix rows \\
Guarded generation completions & 540 \\
Memory guard triggers during MawForge generation & 0 \\
\bottomrule
\end{tabular}
\end{table*}

The three static rejections were Gemma Q8 at 32K with 50\% cache, Gemma Q8 at 32K with 65\% cache, and Qwen Q4 at 32K with 90\% cache.

\begin{table*}[t]
\centering
\scriptsize
\setlength{\tabcolsep}{3pt}
\begin{tabular}{@{}lrrrl@{}}
\toprule
Profile & Context & Cache & Static minimum & Decision \\
\midrule
Gemma Q8 & 32K & 50\% & 20.2782 GiB & rejected \\
Gemma Q8 & 32K & 65\% & 23.6681 GiB & rejected \\
Qwen Q4 & 32K & 90\% & 18.5217 GiB & rejected \\
\bottomrule
\end{tabular}
\end{table*}

\subsection{Measurements}\label{measurements}

The primary dependent variables were decode tokens per second, TTFT, cache hit rate, materialized expert bytes, expert-cache bytes, resident common bytes, KV-cache bytes, process-tree RSS, sampled system-used ratio, and guard status. The CLI telemetry line was persisted with raw request and response bodies, health metrics, parsed JSON telemetry, memory samples, stdout and stderr, and run summaries.

The current worker logs did not collect all preferred macOS memory fields directly. In particular, \passthrough{\lstinline!system\_ram!}, \passthrough{\lstinline!tracked\_metal!}, and \passthrough{\lstinline!metal\_budget!} were not populated by the worker in the latest local logs. The benchmark wrapper collected separate host and process samples. The paper therefore treats process RSS and wrapper samples as memory signals with explicitly named sources rather than merging them into a single unified-memory field.

The reported statistics are descriptive. Each feasible model/context/cache/prompt-class cell contains five valid repetitions, and the source artifacts report means and standard deviations for those cells. This article aggregates selected cells to answer the research questions and does not claim formal hypothesis-test significance. That choice is deliberate: the strongest claim supported by the current matrix is operating-envelope characterization under controlled local conditions, not population-level inference across hardware, prompts, or model distributions.

\subsection{Validity Rules}\label{validity-rules}

A generation row was valid only if the server started, returned HTTP 200 through the OpenAI-compatible endpoint, produced nonzero completion tokens, emitted final telemetry, stayed below the guard threshold, and shut down after the run. A static rejection row was valid when \passthrough{\lstinline!serve plan!} rejected the model/context/cache point before runtime load because the estimated lower bound exceeded the budget.

\section{Results}\label{results}

Unless otherwise noted, numeric results in this section are taken from the benchmark ledger~\cite{mawforgeBenchmarkLedger} and its referenced run artifacts. The validation protocol~\cite{mawforgeValidationProtocol} defines the matrix dimensions and acceptance rules, and the telemetry schema~\cite{mawforgeTelemetrySchema} defines field meanings and derived metrics.

\subsection{RQ1: Feasibility Under a Bounded Budget}\label{rq1-feasibility-under-a-bounded-budget}

MawForge completed all 540 statically feasible generation rows under the 18 GiB target, with no memory-guard triggers during MawForge generation. This supports the feasibility claim for the tested split-pack configurations: MawForge can serve the evaluated GGUF MoE profiles on the 24 GB unified-memory machine when the static planner accepts the row.

The result is strongest when interpreted with the static rejections. MawForge did not force every model/cache/context combination to run. It rejected combinations where common tensors, expert cache, and KV cache already exceeded the target. This is the intended behavior for a bounded local serving system.

\subsection{RQ2: Expert Cache Is Not Monotonic}\label{rq2-expert-cache-is-not-monotonic}

Across the matrix, larger expert caches generally increased hit rate and reduced materialized expert bytes. Throughput did not follow monotonically. The most visible example is Qwen Q8, where all prompt classes at both context lengths favored the smallest tested cache, 15\%.

\begin{table*}[t]
\centering
\scriptsize
\setlength{\tabcolsep}{3pt}
\begin{tabular}{@{}llrrrl@{}}
\toprule
Qwen Q8 context & Prompt & 15\% decode & Highest-cache decode & Highest-cache hit rate & Interpretation \\
\midrule
4K & Long-form writing & 8.51 tok/s & 1.36 tok/s at 45\% & 86.41\% & hit rate improved, decode collapsed \\
4K & Code & 12.09 tok/s & 1.83 tok/s at 45\% & 89.90\% & high cache became a latency failure \\
4K & Summarization & 11.66 tok/s & 1.27 tok/s at 45\% & 89.36\% & TTFT rose to 166.9 s \\
4K & Reasoning & 11.40 tok/s & 1.69 tok/s at 45\% & 89.89\% & smallest cache remained fastest \\
32K & Long-form writing & 11.91 tok/s & 1.41 tok/s at 45\% & 86.41\% & same curve at larger context \\
32K & Code & 12.12 tok/s & 1.56 tok/s at 45\% & 89.90\% & same curve at larger context \\
32K & Summarization & 10.93 tok/s & 1.14 tok/s at 45\% & 89.36\% & high-cache TTFT reached 169.7 s \\
32K & Reasoning & 11.42 tok/s & 1.52 tok/s at 45\% & 89.89\% & high cache minimized I/O, not time \\
\bottomrule
\end{tabular}
\end{table*}

The Qwen Q8 rows show why cache hit rate cannot be used as the objective function. At 32K code, 45\% cache reduced materialized expert I/O to 33.00 GiB and reached 89.90\% hit rate, but decode fell to 1.56 tok/s and TTFT averaged 127.9 s. The row is feasible but not interactive.

Gemma Q8 shows a different curve. At 4K, the best combined cache point was 35\%, not 15\%, but high cache still failed as a throughput setting.

\begin{table*}[t]
\centering
\scriptsize
\setlength{\tabcolsep}{3pt}
\begin{tabular}{@{}rrrrrrrr@{}}
\toprule
Gemma Q8 4K cache & Runs & Decode mean & TTFT mean & Hit rate & Materialized & Max RSS & Max system used \\
\midrule
15\% & 20 & 8.5057 tok/s & 25.9 s & 61.46\% & 165.8767 GiB & 9.2174 GiB & 65.41\% \\
25\% & 20 & 8.6229 tok/s & 24.9 s & 78.83\% & 91.2989 GiB & 11.3355 GiB & 65.08\% \\
35\% & 20 & 13.8591 tok/s & 17.4 s & 86.73\% & 56.8783 GiB & 12.8817 GiB & 67.15\% \\
50\% & 20 & 2.3453 tok/s & 91.6 s & 92.85\% & 29.9465 GiB & 13.9882 GiB & 83.68\% \\
65\% & 20 & 1.3033 tok/s & 116.0 s & 95.03\% & 20.1301 GiB & 15.8544 GiB & 84.09\% \\
\bottomrule
\end{tabular}
\end{table*}

Moving Gemma Q8 from 35\% to 65\% improved hit rate from 86.73\% to 95.03\% and reduced mean materialized I/O from 56.8783 GiB to 20.1301 GiB. Decode fell from 13.8591 tok/s to 1.3033 tok/s. This is the clearest demonstration that lower materialization is not sufficient when resident cache footprint moves the system into a worse operating regime.

\subsection{RQ3: Quantization and Context Change the Operating Point}\label{rq3-quantization-and-context-change-the-operating-point}

Qwen Q4 and Qwen Q8 share the same model family and topology but differ in quantization and footprint. Their cache behavior differs substantially.

At 4K, Qwen Q4 had prompt-dependent optima. Long-form writing favored 15\%, code and reasoning favored 25\%, and summarization had the highest mean at 35\% but better stability at 25\%. At 32K, Qwen Q4 shifted upward: 35\% was best for long-form writing, code, and summarization, while reasoning remained best at 25\%. The 90\% 32K cache point was rejected by static planning because the lower-bound footprint was 18.5217 GiB, above the 18 GiB target.

\begin{table*}[t]
\centering
\scriptsize
\setlength{\tabcolsep}{3pt}
\begin{tabular}{@{}llrrl@{}}
\toprule
Qwen Q4 context & Prompt & Best cache & Best mean decode & Notes \\
\midrule
4K & Long-form writing & 15\% & 18.53 tok/s & smallest cache won \\
4K & Code & 25\% & 14.52 tok/s & 65\% and 90\% regressed \\
4K & Summarization & 35\% & 16.31 tok/s & 25\% was more stable \\
4K & Reasoning & 25\% & 19.87 tok/s & fastest Qwen Q4 4K cell \\
32K & Long-form writing & 35\% & 17.20 tok/s & 25\% nearly tied \\
32K & Code & 35\% & 17.99 tok/s & 65\% fell to 7.87 tok/s \\
32K & Summarization & 35\% & 18.07 tok/s & 65\% fell to 6.75 tok/s \\
32K & Reasoning & 25\% & 18.37 tok/s & strongest Qwen Q4 32K cell \\
\bottomrule
\end{tabular}
\end{table*}

Gemma Q8 preserved a broad 35\% optimum at both context lengths among feasible points. At 32K, 50\% and 65\% were static budget rejections rather than slow-but-loadable rows. The 32K KV cache was 6.8750 GiB, yet 35\% remained feasible and fastest.

\begin{table*}[t]
\centering
\scriptsize
\setlength{\tabcolsep}{3pt}
\begin{tabular}{@{}lrrrrrrr@{}}
\toprule
Gemma Q8 context & Cache & Runs & Decode mean & TTFT mean & Hit rate & Materialized & KV \\
\midrule
4K & 35\% & 20 & 13.8591 tok/s & 17.4 s & 86.73\% & 56.8783 GiB & 0.8594 GiB \\
32K & 35\% & 20 & 14.6371 tok/s & 16.8 s & 86.73\% & 56.8783 GiB & 6.8750 GiB \\
\bottomrule
\end{tabular}
\end{table*}

The context result should not be overgeneralized. The 32K rows used the same 96-token completion limit, so they stress allocation and KV footprint more than long-session behavior. Still, they show that a larger KV cache does not necessarily force the best expert-cache point downward when the static budget remains feasible.

\subsection{RQ4: Speculative Decoding Did Not Improve the Tested MawForge Path}\label{rq4-speculative-decoding-did-not-improve-the-tested-mawforge-path}

The speculative addendum evaluated Gemma Q8 with an MTP draft model. The most focused run used horizon 4 and an acceptance threshold of 0.5 at 25\% expert cache. It accepted 4 of 7 drafted tokens, or 57.1\%, but decoded at 8.31 tok/s compared with a 25\% non-speculative exploratory baseline of 9.52 tok/s. Materialized expert bytes increased from 48.06 GiB to 116.66 GiB.

\begin{table*}[t]
\centering
\scriptsize
\setlength{\tabcolsep}{3pt}
\begin{tabular}{@{}lrrrrrr@{}}
\toprule
Mode & Cache & Decode & TTFT & Accepted & Hit rate & Materialized \\
\midrule
Non-speculative exploratory baseline & 25\% & 9.52 tok/s & 9.8 s & n/a & 78.46\% & 48.06 GiB \\
MTP horizon 4, threshold 0.5 & 25\% & 8.31 tok/s & 14.7 s & 4/7 & 48.38\% & 116.66 GiB \\
\bottomrule
\end{tabular}
\end{table*}

This addendum is not as strong as the 600-row non-speculative matrix because it is a focused comparison rather than a completed factorial study. Its value is diagnostic: speculative acceptance ratio alone was not predictive of speedup. The likely mechanism is route churn. A speculative verification batch can touch a wider set of experts than a single-token decode step, and accepted draft tokens must save enough target work to offset that added materialization.

\subsection{Direct Non-MawForge Load Addendum}\label{direct-non-mawforge-load-addendum}

A guarded direct load attempted to run the original Gemma Q8 GGUF through a direct llama.cpp full-GGUF path at 32K context without MawForge split packs or expert cache. A watchdog terminated the process when system-used memory reached 99.19\%, above the configured 98\% guard. No route output was produced.

\begin{table*}[t]
\centering
\scriptsize
\setlength{\tabcolsep}{3pt}
\begin{tabular}{@{}lrlllrr@{}}
\toprule
Model & Context & Mode & Status & Guard reason & Max system used & Max process-tree RSS \\
\midrule
Gemma 4 26B A4B Q8\_0 & 32K & direct full-GGUF llama.cpp load & memory guard & system\_used\_ratio 0.9919 exceeded 0.9800 & 99.19\% & 16.3313 GiB \\
\bottomrule
\end{tabular}
\end{table*}

This is a load-rejection baseline, not a throughput baseline. It supports the narrower claim that the tested direct full-GGUF Metal load at 32K was unsafe or impractical on this 24 GB machine without an external guard. It does not prove that all llama.cpp configurations, lower contexts, lower quantization levels, CPU-only modes, or alternate offload policies would fail.

\section{Discussion}\label{discussion}

\subsection{Effectiveness of MawForge}\label{effectiveness-of-mawforge}

Within the tested boundary, MawForge is effective at what it is designed to do: enforce a memory budget, reject over-budget cache/context settings before load, and serve large local MoE models by materializing routed experts on demand. The completed matrix gives a stronger claim than an isolated demonstration because it covers all statically feasible cells in the declared factorial protocol.

The system should not be described as a universal speedup over conventional local inference. The strongest evidence is a same-machine feasibility and operating-envelope result. MawForge turns configurations that may be unsafe or impractical under full residency into measurable bounded configurations, but performance depends on a tuned cache point.

\subsection{Cache Hit Rate Is an Incomplete Objective}\label{cache-hit-rate-is-an-incomplete-objective}

The most important empirical pattern is that hit rate and throughput diverge. In a pure cache model, one might expect larger caches to improve latency because they reduce misses. On unified-memory Apple Silicon, cache footprint itself has a cost. Once the footprint approaches a pressure region, reduced materialization can be outweighed by memory compression, paging, lower effective bandwidth, or allocator behavior.

This implies that MawForge needs an optimizer that minimizes end-to-end latency subject to a memory budget, not a policy that maximizes cache hit rate. A practical tuner should use decode tok/s, TTFT, materialized bytes, cache hit rate, KV bytes, RSS, wrapper memory samples, and static feasibility together.

\subsection{Quantization Is a Systems Variable}\label{quantization-is-a-systems-variable}

Quantization changed both footprint and optimum. Qwen Q4 was generally faster than Qwen Q8 and had a broader feasible cache region, but it did not produce a universal cache recommendation. Qwen Q4 at 4K favored different cache settings for different prompt classes; at 32K, 35\% became the broad default while reasoning favored 25\%. Quantization should therefore be treated as part of the serving configuration, not only as a model-quality or storage decision.

\subsection{Context Length Has Two Roles}\label{context-length-has-two-roles}

Context length increases KV-cache footprint and can move static feasibility boundaries. Gemma Q8 at 32K rejected 50\% and 65\% cache settings that were feasible at 4K. However, context length did not always reduce the best feasible cache point. Gemma Q8 still favored 35\% at 32K, and Qwen Q4 moved upward from a 25\% conservative 4K default to a 35\% broad 32K default. The interaction depends on model topology, KV size, quantization, and prompt route locality.

\subsection{Speculation Requires MoE-Aware Accounting}\label{speculation-requires-moe-aware-accounting}

The speculative addendum argues against importing dense-model speculation expectations without measurement. Accepted draft tokens matter, but expert materialization can erase the benefit. A future speculative study should record per-step drafted tokens, accepted tokens, target verification width, expert misses, materialized bytes, slot evictions, draft-model memory, and draft KV memory.

\section{Threats to Validity}\label{threats-to-validity}

The evaluation is limited to one 24 GB unified-memory Apple Silicon machine. Results may differ across memory sizes, SSD performance, macOS versions, thermal states, power modes, and CPU/GPU scheduling behavior.

The prompt suite contains four deterministic prompt classes and 96-token completions. This supports controlled cache comparisons but does not characterize long interactive sessions, multi-turn state growth, streaming workloads, or multi-user serving.

The matrix evaluates MawForge\textquotesingle s non-speculative path thoroughly for the declared profiles, but direct non-MawForge comparison is represented by one guarded Gemma Q8 32K load attempt. Broader baselines are needed for lower contexts, lower quantization, CPU-only settings, alternate offload policies, and conventional llama.cpp configurations that use different memory behavior.

The evidence artifacts were collected locally by the MawForge project rather than by an independent laboratory. The artifacts are sufficient for internal reproducibility and paper inspection, but external replication remains necessary before generalizing the measured operating points beyond the tested environment.

The current worker telemetry does not directly populate every macOS memory-pressure field. Wrapper samples provide useful independent signals, but timestamp-aligned OS pressure, Metal allocation tracking, and page-compression counters would strengthen causal interpretation of high-cache regressions.

The speculative result is diagnostic rather than comprehensive. It should not be generalized to all MTP settings, all MoE models, or all draft models without a larger factorial speculative study.

Finally, the evaluation measures systems behavior, not answer quality. Temperature was fixed and completions were constrained for reproducibility; the paper does not claim that one cache setting improves semantic quality over another.

\section{Artifact Availability and Reproducibility}\label{artifact-availability-and-reproducibility}

The local evidence base is stored under \passthrough{\lstinline!docs/evidence!}. The benchmark ledger is \passthrough{\lstinline!docs/evidence/benchmark-results.md!}. The validation protocol is \passthrough{\lstinline!docs/evidence/validation-protocol.md!}. The telemetry schema is \passthrough{\lstinline!docs/evidence/telemetry-schema.md!}. The static manifest is \passthrough{\lstinline!docs/evidence/runs/20260608-203646Z-static-plan/manifest.json!}, with matrix rows in \passthrough{\lstinline!matrix.jsonl!} and static planning rows in \passthrough{\lstinline!plans.jsonl!}.

The generation artifacts include raw requests, responses, parsed telemetry, memory samples, and run summaries under \passthrough{\lstinline!docs/evidence/runs!}. The direct non-MawForge addendum is documented in \passthrough{\lstinline!docs/evidence/direct-non-mawforge-load-attempt.md!} and \passthrough{\lstinline!docs/evidence/runs/20260609-062219Z-direct-llama!}.

The primary static planning command is:

\begin{lstlisting}[language=sh]
scripts/bench/mawforge_matrix_runner.py
\end{lstlisting}

The guarded generation command template is:

\begin{lstlisting}[language=sh]
scripts/bench/mawforge_generation_runner.py \
  --matrix-path docs/evidence/runs/20260608-203646Z-static-plan/matrix.jsonl
\end{lstlisting}

Each served row uses the same core MawForge command shape:

\begin{lstlisting}[language=sh]
target/release/mawforge serve \
  --pack "$PACK" \
  --model-id "$MODEL_ID" \
  --host 127.0.0.1 \
  --port "$PORT" \
  --memory-budget-gib 18 \
  --context "$CONTEXT" \
  --expert-cache-percent "$CACHE_PERCENT" \
  --max-tokens-default 96 \
  --no-speculative
\end{lstlisting}

\section{Answers to the Research Questions}\label{answers-to-the-research-questions}

\textbf{RQ1.} Yes, for the tested statically feasible configurations. MawForge completed 540 guarded generation rows under an 18 GiB target and rejected 60 over-budget matrix rows through static planning. This supports bounded local serving feasibility for the evaluated profiles on the tested 24 GB machine.

\textbf{RQ2.} Expert cache percentage has a non-monotonic relationship with throughput. Larger caches improved hit rate and reduced materialized bytes, but high-cache rows often became latency failures. Qwen Q8 favored 15\% at both 4K and 32K across all prompt classes. Gemma Q8 favored 35\% and then collapsed at 50\% and 65\% at 4K despite higher hit rates.

\textbf{RQ3.} Quantization and context changed the operating point. Qwen Q4 had prompt-dependent 4K optima and a broader 35\% 32K default, while Qwen Q8 consistently favored 15\%. Gemma Q8 preserved a 35\% optimum at 32K among feasible points, but higher 32K cache settings were rejected by static planning.

\textbf{RQ4.} The tested speculative path did not improve throughput. Gemma MTP horizon 4 accepted 4 of 7 draft tokens, but decode fell from 9.52 tok/s to 8.31 tok/s and materialized expert bytes rose from 48.06 GiB to 116.66 GiB.

\textbf{RQ5.} The completed evidence supports bounded MawForge feasibility, non-monotonic cache behavior, model-specific cache tuning, static rejection as a safety mechanism, and a narrow direct-load rejection comparison. It does not support universal cache recommendations, universal speculative speedup claims, broad direct-baseline superiority, or answer-quality claims.

\section{Conclusion}\label{conclusion}

MawForge demonstrates that local sparse-expert inference can be implemented as bounded expert materialization rather than full expert residency. On the tested 24 GB unified-memory machine, the system served every statically feasible row in a 600-row non-speculative validation matrix under an 18 GiB target and rejected over-budget cells before runtime load. This is meaningful because the tested model files are substantially larger than the comfortable resident-memory envelope of the machine.

The evidence also narrows the correct performance claim. MawForge is not a cache-maximization system. It is a memory-bounded execution system whose performance depends on finding the point where expert reuse improves enough to justify resident cache footprint. Qwen Q8, Qwen Q4, and Gemma Q8 occupy different points in this design space. High cache can increase hit rate while making the system slower, and speculative decoding can increase accepted draft tokens while increasing expert materialization.

The practical conclusion is that MawForge makes constrained local MoE serving measurable and controllable. Its strongest current result is not a single throughput number but a validated operating-envelope method: plan the budget, execute only feasible cells, collect route-aware telemetry, and choose cache settings by measured latency and materialization under host memory pressure.

\section*{AI Assistance Disclosure}

The author used OpenAI Codex 5.5 and Anthropic Claude Opus 4.8 as text-to-text generative AI tools during manuscript preparation. These tools were used to assist with drafting, editing, LaTeX formatting, and submission compliance review. The author reviewed, edited, and verified the manuscript, including all technical claims, citations, data, and conclusions, and accepts full responsibility for the submitted work.

\end{document}